%% file: ms.tex
\documentclass[conference]{IEEEtran}

\usepackage{amsmath}
\usepackage{amsfonts}
\usepackage{graphicx}
\usepackage{algorithmic}
\usepackage{algorithm}
\usepackage{color}

\usepackage[keeplastbox]{flushend}

\usepackage{import}
\usepackage{multirow}
\usepackage{cite}
\usepackage[export]{adjustbox}
\usepackage{xspace}
\usepackage{balance}

\newcommand{\fig}[1]{Fig.~\ref{#1}}

\newcommand{\secref}[1]{Section~\ref{#1}}

\renewcommand{\baselinestretch}{0.99}
\newcommand{\steps}{h}

\title{Machine Learning Approaches to Energy Consumption Forecasting in Households}

\author{
    \IEEEauthorblockN{Riccardo Bonetto, Michele Rossi}
    \IEEEauthorblockA{Department of Information Engineering (DEI)\\
     University of Padova, Via G. Gradenigo 6/B, 35131 Padova (PD), Italy
    \\\{\tt {bonettor, rossi\}@dei.unipd.it}}
}

\begin{document}
\maketitle

\algsetup{
	linenosize=\small,
	linenodelimiter=.
}
\input{"abstract.tex"}
\input{"intro.tex"}

\input{"SVM.tex"}

\input{"simSetup.tex"}
\input{"results.tex"}
\input{"conclusions.tex"}
\bibliographystyle{IEEEtran}
\bibliography{bibliography}


\end{document}

%% file: abstract.tex

\begin{abstract}
We consider the problem of power demand forecasting in residential micro-grids. Several approaches using ARMA models, support vector machines, and recurrent neural networks that perform one-step ahead predictions have been proposed in the literature. Here, we extend them to perform multi-step ahead forecasting and we compare their performance. Toward this end, we implement a parallel and efficient training framework, using power demand traces from real deployments to gauge the accuracy of the considered techniques. Our results indicate that machine learning schemes achieve smaller prediction errors in the mean and the variance with respect to ARMA, but there is no clear algorithm of choice among them. Pros and cons of these approaches are discussed and the solution of choice is found to depend on the specific use case requirements. A hybrid approach, that is driven by the prediction interval, the target error, and its uncertainty, is then recommended.
\end{abstract}

%% file: intro.tex

\section{Introduction}\label{sec:intro}

In the last few years, the raising concern for greenhouse gas emissions, the growth in the electrical power demand, the diffusion of domestic generation plants based of renewables, and the integration of sensing and metering devices into power distribution grids has led to the deployment of several smart grids around the world. As noted in~\cite{art_1}, they are one of the key enablers for the development of smart cities.

As for smart grid technology, a great effort has been devoted to developing distributed control techniques that boost the efficiency of electrical grids in the presence of end users with power generation capabilities (prosumers), see for instance~\cite{TokenRing,tii,seville}. Moreover, demand response policies that influence the energy consumption profile of the prosumers providing economic and power efficiency benefits are being investigated~\cite{dr_1}. 

Efficient power consumption forecasting algorithms can provide further benefits to the smart grid control process. For example, in~\cite{add_1,add_2} forecasting is utilized to assess what fraction of the generated power has to be locally stored for later use and what fraction of it can instead be fed to the loads or injected into the grid. Moreover, in~\cite{seville} prosumers' power generation and consumption forecasts are used to determine the amount of energy that has to be injected in an isolated power grid to stabilize the aggregated power consumption.

Lately, several techniques have been developed and an increasing attention is being paid to machine learning approaches such Artificial Neural Networks (ANN)~\cite{add_5} and Support Vector Machines (SVM)~\cite{svm}. These methods, however, are known to be computationally intensive~\cite{svm_complexity,ann_complexity}. For this reason, lightweight forecasting solutions are much needed to utilize them in prosumers' installations featuring off-the-shelf computing hardware.

In recent work, ANNs and SVMs have been successfully (and increasingly) exploited to forecast power consumption data, see for example~\cite{for_1,svm_1}. In this paper, we provide a comparison between four different forecasting methods. Each of these can be executed by off-the-shelf computing hardware and can perform day-ahead and multi-step ahead predictions, i.e., when the output is a vector of forecast power demands into the future.

The first technique that we consider is an Auto-Regressive Moving Average (ARMA) model, whose results are used as a baseline to quantify the forecast accuracy gain provided by machine learning algorithms. The second method that we investigate is based on $\steps$ SVMs which are trained and executed in a parallel fashion, where $\steps$ is the number of time steps into the future to be forecasted. The last two methods employ ANNs. The third one is based on a Nonlinear Auto-Regressive (NAR) recurrent ANN, while the fourth features a Long Short-Term Memory (LSTM) ANN. For each of the considered ANN approaches, a single network topology is defined and is then trained $\steps$ times (one training per time step). The weights and biases for each of the $\steps$ training phases are then utilized to generate an $\steps$-steps ahead forecast vector. This approach allows performing the training and forecast processes in a parallel fashion and to only store $2\steps$ matrices.

The main contribution of the present work consists of carrying out a performance comparison of machine learning solutions from the state-of-the-art, which are seldom compared against one another. Besides, we also extend our comparison to LSTM neural networks, which to the best of our knowledge were never used for energy demand forecasting in smart grids.

The rest of this work is structured as follows. In \secref{ml} we briefly introduce the three considered machine learning forecasting techniques, namely, support vector machines, non linear auto-regressive neural networks, and long short-term memory neural networks. In \secref{simSetup} we discuss a parallel framework that reduces the computational time required by the training phase and makes it possible to implement the considered forecasting solutions in off-the-shelf computing devices. Moreover, we describe the experimental setup that we used to assess its performance and in \secref{results} we test machine learning forecasting approaches against an ARMA model. Finally, in \secref{conclusions} we draw our conclusions.

%% file: SVM.tex

\section{Machine Learning Techniques for Forecasting}\label{ml}

In this section, we briefly describe the considered machine learning techniques, along with the adopted forecasting architectures.

\subsection{Support Vector Machines (SVM)}\label{svm}

SVMs~\cite{svm} can be used for classification and regression tasks. When used for regression, the common approach is called ``\mbox{$\epsilon$-insensitive} support vector regression''~\cite{svm}. Let $h$ be future horizon of the forecast and let $X_{n}^{\tau}$ be the last $n$ values of the time series $X=(x_1,x_2,\dots)$ at the current time step $\tau \geq 1$. Then, this approach seeks a function $f(X_{n}^{\tau})$ such that a suitable distance $||f(X_{n}^{\tau})-x_{\tau+h}||$ is minimized and in any case is no greater than the parameter $\epsilon$. To do so, a convex optimization problem is set up and solved. This guarantees that, if the problem is feasible, the  solution is the best one. In the case where the optimization problem does not have any feasible solution, a tolerance parameter on the $\epsilon$ threshold is introduced. The use of SVMs for regression tasks is appealing since they guarantee that the forecast error is bounded by $\epsilon$.  

\subsection{Nonlinear Auto-Regressive (NAR) Neural Networks}\label{nar}

Non linear auto-regressive neural networks (NARs) are recurrent ANNs performing regression tasks on time series~\cite{NNs}. A NAR network operates on a time series $X$ by processing, at each time step $\tau$, the subsequence $X_{n}^{\tau}$ (i.e., the last $n$ values of $X$, i.e., ($x_{\tau-n+1},\dots,x_\tau$)) and the previous NAR's output $\hat{x}_{\tau}$. The parameter $n$ determines the ANN's memory, i.e., how far in the past the NAR is required to track the correlation structure of the input data. In order to capture the nonlinear structure of the considered time series, it is required that neurons in the hidden layers have nonlinear activation functions. The output layer, instead, is composed of neurons with linear activation functions, so that the output is not bounded to any particular dimension. 

\begin{figure}
\center
\def\svgwidth{0.7\columnwidth}
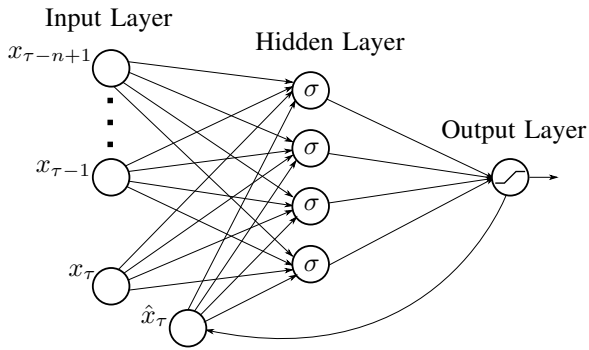
\caption{Example of NAR network with $n$ inputs, one hidden layer with $4$ neurons and $1$ output neuron.\label{fig:nar}}
\end{figure}

\fig{fig:nar} shows an example NAR network: it takes as input the sequence $X_{n}^{\tau}$ and the output generated by the same network in the previous time step. These inputs are processed by a hidden layer composed of four neurons with sigmoidal activation functions ($\sigma$ in the figure). The output of the hidden layer is processed by a linear neuron to produce the desired result.

\subsection{Long Short-Term Memory (LSTM) Neural Networks}\label{lstm}

Long Short-Term Memory (LSTM) networks~\cite{lstm} are a particular class of recurrent ANNs where the neurons in the hidden layers are replaced by the so-called \textit{memory cells} (MC). A MC is a particular structure that allows storing or forgetting information about past network states. This is made possible by structures called \textit{gates}. Gates are composed of a cascade of a neuron with sigmoidal activation function and a pointwise multiplication block. 

\begin{figure}
\center
\def\svgwidth{0.7\columnwidth}
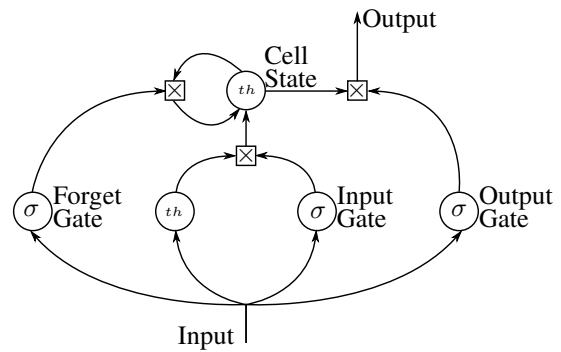
\caption{Example of LSTM Memory Cell (MC).\label{fig:lstm_cell}}
\end{figure}

\fig{fig:lstm_cell} shows a typical MC structure. The input gate is a neuron with sigmoidal activation function ($\sigma$). Its output determines the fraction of MC input that is fed to the cell state block. Similarly, the forget gate processes the information that is recurrently fed back into the cell state block. The output gate, instead, determines the fraction of the output of the cell state that is to be used as output of the MC at each time step. The gates' neurons usually have sigmoidal activation functions ($\sigma$), while the input and cell state use the hyperbolic tangent ($th$) activation function. All the internal connections of the MC have unitary weight. Thanks to this architecture, the output of each memory cell possibly depends on the entire sequence of past states. This make LSTM ANNs particularly suitable for processing time series with long time dependencies (i.e., inter-sample correlation).

\begin{figure}
\center
\def\svgwidth{0.7\columnwidth}
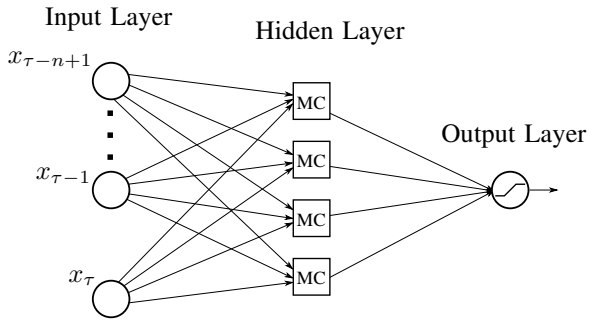
\caption{Example of LSTM network with $n$ inputs, one hidden layer with $4$ Memory Cells (MC) and $1$ output neuron.\label{fig:lstm}}
\end{figure}

\fig{fig:lstm} shows an example of the LSTM ANN architectures that we used in this work. As done with the NAR networks of \secref{nar}, we consider the sequence $X_{n}^{\tau}$ as the network input vector. These $n$ time samples are fed as input to the memory cells ($4$ MC cells are shown in \fig{fig:lstm}). As time ($\tau$) advances, the output of the memory cells depends on the current input sequence $X_{n}^{\tau}$ and as well on the previous ones $X_{n-i}^{\tau-i},\ i=1,\dots,\tau-n$. As with the NAR network, the output of the hidden layer is filtered through a linear neuron to obtain the network output.

%% file: figures/nar.eps_tex
\begingroup%
  \makeatletter%
  \providecommand\color[2][]{%
    \errmessage{(Inkscape) Color is used for the text in Inkscape, but the package 'color.sty' is not loaded}%
    \renewcommand\color[2][]{}%
  }%
  \providecommand\transparent[1]{%
    \errmessage{(Inkscape) Transparency is used (non-zero) for the text in Inkscape, but the package 'transparent.sty' is not loaded}%
    \renewcommand\transparent[1]{}%
  }%
  \providecommand\rotatebox[2]{#2}%
  \ifx\svgwidth\undefined%
    \setlength{\unitlength}{816.28798828bp}%
    \ifx\svgscale\undefined%
      \relax%
    \else%
      \setlength{\unitlength}{\unitlength * \real{\svgscale}}%
    \fi%
  \else%
    \setlength{\unitlength}{\svgwidth}%
  \fi%
  \global\let\svgwidth\undefined%
  \global\let\svgscale\undefined%
  \makeatother%
  \begin{picture}(1,0.72924702)%
    \put(0,0){\includegraphics[width=\unitlength]{./figures/nar.eps}}%
    \put(-0.1,0.69){\color[rgb]{0,0,0}\makebox(0,0)[lb]{\smash{Input Layer}}}%
    \put(0.35085657,0.64){\color[rgb]{0,0,0}\makebox(0,0)[lb]{\smash{Hidden Layer}}}%
    \put(0.75,0.45){\color[rgb]{0,0,0}\makebox(0,0)[lb]{\smash{Output Layer}}}%
    \put(0.455,0.54){\color[rgb]{0,0,0}\makebox(0,0)[lb]{\smash{$\sigma$}}}%
    \put(0.455,0.415){\color[rgb]{0,0,0}\makebox(0,0)[lb]{\smash{$\sigma$}}}%
    \put(0.455,0.295){\color[rgb]{0,0,0}\makebox(0,0)[lb]{\smash{$\sigma$}}}%
    \put(0.455,0.165){\color[rgb]{0,0,0}\makebox(0,0)[lb]{\smash{$\sigma$}}}%
    \put(-0.18,0.62302942){\color[rgb]{0,0,0}\makebox(0,0)[lb]{\smash{$x_{\tau-n+1}$}}}%
    \put(-0.12,0.37661777){\color[rgb]{0,0,0}\makebox(0,0)[lb]{\smash{$x_{\tau-1}$}}}%
    \put(-0.05,0.155){\color[rgb]{0,0,0}\makebox(0,0)[lb]{\smash{$x_{\tau}$}}}%
    \put(0.105,0.054){\color[rgb]{0,0,0}\makebox(0,0)[lb]{\smash{$\hat{x}_{\tau}$}}}%
  \end{picture}%
\endgroup%

%% file: figures/lstm_cell.eps_tex
\begingroup%
  \makeatletter%
  \providecommand\color[2][]{%
    \errmessage{(Inkscape) Color is used for the text in Inkscape, but the package 'color.sty' is not loaded}%
    \renewcommand\color[2][]{}%
  }%
  \providecommand\transparent[1]{%
    \errmessage{(Inkscape) Transparency is used (non-zero) for the text in Inkscape, but the package 'transparent.sty' is not loaded}%
    \renewcommand\transparent[1]{}%
  }%
  \providecommand\rotatebox[2]{#2}%
  \ifx\svgwidth\undefined%
    \setlength{\unitlength}{892.92587891bp}%
    \ifx\svgscale\undefined%
      \relax%
    \else%
      \setlength{\unitlength}{\unitlength * \real{\svgscale}}%
    \fi%
  \else%
    \setlength{\unitlength}{\svgwidth}%
  \fi%
  \global\let\svgwidth\undefined%
  \global\let\svgscale\undefined%
  \makeatother%
  \begin{picture}(1,0.61898122)%
    \put(0,0){\includegraphics[width=\unitlength]{./figures/lstm_cell.eps}}%
    \put(0.086,0.3){\color[rgb]{0,0,0}\makebox(0,0)[lb]{\smash{Forget}}}%
    \put(0.086,0.25){\color[rgb]{0,0,0}\makebox(0,0)[lb]{\smash{Gate}}}%
    \put(0.35408475,0.){\color[rgb]{0,0,0}\makebox(0,0)[lb]{\smash{Input}}}%
    \put(0.695,0.3){\color[rgb]{0,0,0}\makebox(0,0)[lb]{\smash{Input}}}%
    \put(0.695,0.25){\color[rgb]{0,0,0}\makebox(0,0)[lb]{\smash{Gate}}}%
    \put(0.75,0.68){\color[rgb]{0,0,0}\makebox(0,0)[lb]{\smash{Output}}}%
    \put(1,0.3){\color[rgb]{0,0,0}\makebox(0,0)[lb]{\smash{Output}}}%
    \put(1,0.25){\color[rgb]{0,0,0}\makebox(0,0)[lb]{\smash{Gate}}}%
    \put(0.54,0.6){\color[rgb]{0,0,0}\makebox(0,0)[lb]{\smash{Cell}}}%
    \put(0.54,0.55){\color[rgb]{0,0,0}\makebox(0,0)[lb]{\smash{State}}}%
    \put(0.3255,0.529){\color[rgb]{0,0,0}\makebox(0,0)[lb]{\smash{$\times$}}}%
    \put(0.716,0.529){\color[rgb]{0,0,0}\makebox(0,0)[lb]{\smash{$\times$}}}%
    \put(0.478,0.388){\color[rgb]{0,0,0}\makebox(0,0)[lb]{\smash{$\times$}}}%
    \put(0.482,0.536){\color[rgb]{0,0,0}\makebox(0,0)[lb]{\smash{\tiny{$th$}}}}%
    \put(0.326,0.273){\color[rgb]{0,0,0}\makebox(0,0)[lb]{\smash{\tiny{$th$}}}}%
    \put(0.637,0.27){\color[rgb]{0,0,0}\makebox(0,0)[lb]{\smash{$\sigma$}}}%
    \put(0.94,0.273){\color[rgb]{0,0,0}\makebox(0,0)[lb]{\smash{$\sigma$}}}%
    \put(0.023,0.273){\color[rgb]{0,0,0}\makebox(0,0)[lb]{\smash{$\sigma$}}}%
  \end{picture}%
\endgroup%

%% file: figures/lstm.eps_tex
\begingroup%
  \makeatletter%
  \providecommand\color[2][]{%
    \errmessage{(Inkscape) Color is used for the text in Inkscape, but the package 'color.sty' is not loaded}%
    \renewcommand\color[2][]{}%
  }%
  \providecommand\transparent[1]{%
    \errmessage{(Inkscape) Transparency is used (non-zero) for the text in Inkscape, but the package 'transparent.sty' is not loaded}%
    \renewcommand\transparent[1]{}%
  }%
  \providecommand\rotatebox[2]{#2}%
  \ifx\svgwidth\undefined%
    \setlength{\unitlength}{816.28798828bp}%
    \ifx\svgscale\undefined%
      \relax%
    \else%
      \setlength{\unitlength}{\unitlength * \real{\svgscale}}%
    \fi%
  \else%
    \setlength{\unitlength}{\svgwidth}%
  \fi%
  \global\let\svgwidth\undefined%
  \global\let\svgscale\undefined%
  \makeatother%
  \begin{picture}(1,0.72924702)%
    \put(0,0){\includegraphics[width=\unitlength]{./figures/lstm.eps}}%
    \put(-0.1,0.63){\color[rgb]{0,0,0}\makebox(0,0)[lb]{\smash{Input Layer}}}%
    \put(0.35085657,0.6){\color[rgb]{0,0,0}\makebox(0,0)[lb]{\smash{Hidden Layer}}}%
    \put(0.75,0.38){\color[rgb]{0,0,0}\makebox(0,0)[lb]{\smash{Output Layer}}}%
    \put(0.44,0.45){\color[rgb]{0,0,0}\makebox(0,0)[lb]{\smash{\scriptsize{MC}}}}%
    \put(0.44,0.325){\color[rgb]{0,0,0}\makebox(0,0)[lb]{\smash{\scriptsize{MC}}}}%
    \put(0.44,0.2){\color[rgb]{0,0,0}\makebox(0,0)[lb]{\smash{\scriptsize{MC}}}}%
    \put(0.44,0.075){\color[rgb]{0,0,0}\makebox(0,0)[lb]{\smash{\scriptsize{MC}}}}%
    \put(-0.18,0.55){\color[rgb]{0,0,0}\makebox(0,0)[lb]{\smash{$x_{\tau-n+1}$}}}%
    \put(-0.12,0.3){\color[rgb]{0,0,0}\makebox(0,0)[lb]{\smash{$x_{\tau-1}$}}}%
    \put(-0.05,0.08){\color[rgb]{0,0,0}\makebox(0,0)[lb]{\smash{$x_{\tau}$}}}%
  \end{picture}%
\endgroup%

%% file: simSetup.tex

\section{Experimental Setup}\label{simSetup}

In this section, we describe the framework that we used to perform the power forecasts with off-the-shelf hardware. Moreover, we introduce the experimental setup (power demand traces and configuration parameters for the considered schemes) for the numerical assessments of \secref{results}. 

\subsection{Parallel Framework}

Our parallel approach splits the forecasting problem into embarrassingly parallel subproblems. Each subproblem corresponds to the forecast of one of the values of the \mbox{multi-step} forecast vector. With respect to the SVM approach, for a $\steps$-step ahead forecast, $\steps$ SVMs are trained in parallel exploiting the \mbox{multi-core} architecture of commercial CPUs and GPUs. Each SVM is trained to solve one of the $\steps$ subproblems of the $\steps$-step ahead forecast. Similarly, for the considered ANN architectures a single network topology is trained multiple times. Each training time returns a set of weights and biases corresponding to the solution of one of the $\steps$ subproblems. Upon completion of the parallel training phase, $\steps$-step ahead forecast vectors can be obtained in a parallel fashion as well.

\subsection{Parameters Setup}
In order to assess the performance of the considered forecasting methods, we utilized the dataset in~\cite{UCIDataSet}. It contains active power consumption measurements for a single household. Measurements were taken every $60$ seconds during a period of $4$ years, resulting in more than $2$ million time samples. Part of this dataset has been used as training set for the considered forecasting approaches, while the remaining time samples were used to assess the accuracy of the obtained forecasts.

For all the approaches, we performed $120$-step ahead forecasts ($\steps=120$, one step corresponds to one minute) using sequences of $n=30$ past time samples. Given the time scale of the considered dataset, this corresponds to $2$ \mbox{hour-ahead} forecasts using measurements from the past $30$ minutes. For the ARMA, SVM and NAR approaches we considered a training set of $5,000$ time samples. Our experimental results have shown that increasing the size of the training set beyond this leads to marginal accuracy improvements at the cost of a significantly increased training time. For LSTM, instead, we set the training set size to $100,000$ time samples. This value led to the best accuracy for this scheme and to a still acceptable training time with off-the-shelf hardware. 

The NAR network that we used for the results in the next section is configured as follows:
\begin{itemize}
	\item it takes the subsequence $X_{30}^{\tau}$ as input (i.e., a time horizon of $30$ minutes is used to forecast future values);
	\item it has one hidden layer with $40$ neurons, each of them with sigmoidal activation function;
	\item it has one output neuron with linear activation function.
\end{itemize}
The training algorithm chosen for the NAR approach is the Levenberg-Marquardt with Bayesian weights regularization~\cite{bishop}. This algorithm is particularly suited for time series exhibiting a noisy behavior. 

The LSTM network that we used for the results in the next section is configured as follows:
\begin{itemize}
	\item it takes the subsequence $X_{30}^{\tau}$ as input (i.e., a time horizon of $30$ minutes is used to forecast future values);
	\item it has one hidden layer with $50$ memory cells, each of them with softsign activation function~\cite{softsign};
	\item it has one output neuron with linear activation function.
\end{itemize}

The LSTM network has been trained using the ADAGRAD algorithm~\cite{adagrad}. This training method exhibits an improved convergence rate over standard gradient descent schemes thanks to a dynamically adjustable learning rate.

%% file: results.tex

\section{Results}\label{results}

Next, we present the experimental results obtained through our parallel forecasting framework and the parameters setup of \secref{simSetup}. We compare the performance of the considered machine learning approaches against that of a state-of-the-art ARMA model in terms of mean absolute error and error variance.

\begin{figure}
\includegraphics[width=\columnwidth]{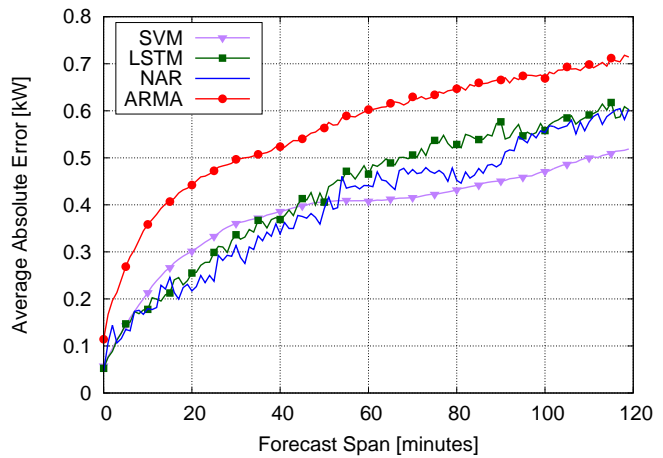}
\caption{Mean absolute error for the estimated ARMA model and the proposed NAR-based forecasting scheme.}\label{fig:error}
\end{figure}

\fig{fig:error} shows the mean absolute error of the forecasts obtained by the ARMA model and those obtained with the SVM, NAR, and LSTM approaches. The mean absolute error has been computed for each forecast as the arithmetic average of the distance between the points estimated by the models and the target values of the input time series. A first consideration is that all the machine learning approaches outperform the results obtained by ARMA. Also, NAR and LSTM have similar average performance. However, LSTM requires a training set that is $20$ times bigger than that used to train the NAR network. The SVM approach in the first $40$ time steps exhibits a slightly higher error with respect to NAR and LSTM. However, for longer time spans it achieves the best forecast accuracy. These results suggest the adoption of a hybrid forecasting approach where the first forecasts are computed through a NAR ANN and the last ones are obtained via SVM. The point that determines the transition between the NAR network and the SVM model depends on the dataset characteristics and should be determined for every use-case.

\begin{figure}
\includegraphics[width=\columnwidth]{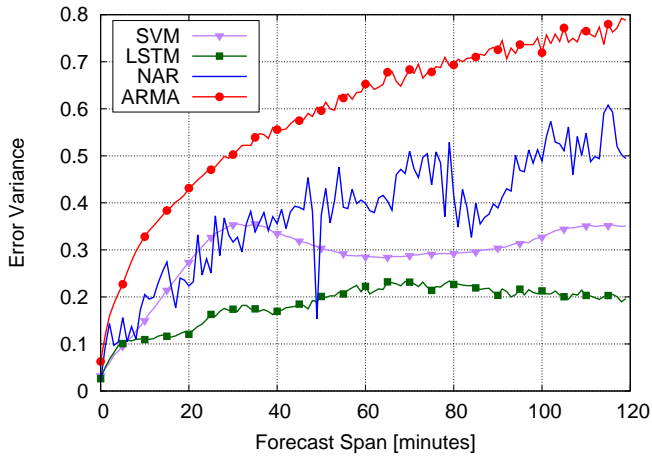}
\caption{Error variance for the estimated ARMA model and the proposed NAR-based forecasting scheme.}\label{fig:variance}
\end{figure}

In \fig{fig:variance}, we show the variance for the prediction errors of \fig{fig:error}, which is related to the prediction uncertainty. As a first result, we note that  for the ARMA and NAR approaches the error variances exhibit the same growth trend as for the average errors of \fig{fig:error}. This confirms that an increasing time window corresponds to a correspondingly increasing uncertainty in the prediction accuracy. Nevertheless, especially in the first $20$ minutes, the ARMA's error variance grows much faster than that of NAR, making the latter a better approach. SVM and LSTM techniques exhibit a considerably lower error variance with respect to ARMA and NAR. The error variance of SVM grows as fast as the NAR's one within the first $40$ prediction steps and then drops, reaching a minimum around $\steps=60$ minutes. SVM resulted to be the algorithm of choice when predicting far ahead in time, as it obtains the smallest error, while also showing the second-smallest uncertainty (error variance). This is due to the SVM parameter $\epsilon$, which sets a bound on the maximum forecasting error. The aforementioned hybrid scheme, i.e., using NAR for short prediction intervals (e.g., up to $40$ minutes) and then switching to SVM is also supported by the result of \fig{fig:variance}. Finally, LSTM exhibits the smallest variance with respect to all other methods. This means that, despite not being the most accurate forecasting scheme, it guarantees that the error fluctuations are small.

%% file: conclusions.tex

\section{Conclusions}\label{conclusions}

In this work we have presented a comparison of the performance of different machine learning approaches in terms of multi-steps ahead forecasting error and error variance. After briefly reviewing machine learning approaches for forecasting from the literature, namely, SVM, NAR and LSTM ANNs, we described their forecasting architectures and the dataset that has been used for their experimental results. The obtained results have been compared to the ones obtained by an ARMA model. All the machine learning approaches outperform ARMA, whereas no single algorithm of choice exists among SVM, NAR and LSTM. The LSTM network exhibits a slightly worse prediction accuracy with respect to the others, but it has the smallest error variance. NAR exhibits the best forecasting accuracy for short prediction windows. Instead, SVM shows a complementary behavior, guaranteeing the best accuracy for the estimation of power demands far ahead in time. Our results suggest the adoption of a hybrid approach, which entails the use of NAR for short time horizons and SVM for long prediction intervals.

%% file: ms.bbl
\begin{thebibliography}{10}
\providecommand{\url}[1]{#1}
\csname url@samestyle\endcsname
\providecommand{\newblock}{\relax}
\providecommand{\bibinfo}[2]{#2}
\providecommand{\BIBentrySTDinterwordspacing}{\spaceskip=0pt\relax}
\providecommand{\BIBentryALTinterwordstretchfactor}{4}
\providecommand{\BIBentryALTinterwordspacing}{\spaceskip=\fontdimen2\font plus
\BIBentryALTinterwordstretchfactor\fontdimen3\font minus
  \fontdimen4\font\relax}
\providecommand{\BIBforeignlanguage}[2]{{%
\expandafter\ifx\csname l@#1\endcsname\relax
\typeout{** WARNING: IEEEtran.bst: No hyphenation pattern has been}%
\typeout{** loaded for the language `#1'. Using the pattern for}%
\typeout{** the default language instead.}%
\else
\language=\csname l@#1\endcsname
\fi
#2}}
\providecommand{\BIBdecl}{\relax}
\BIBdecl

\bibitem{art_1}
K.~{Geisler}, ``{The Relationship Between Smart Grids and Smart Cities},''
  \emph{IEEE Smart Grid Newsletter Compendium}, May 2013.

\bibitem{TokenRing}
P.~{Tenti}, A.~{Costabeber}, P.~{Mattavelli}, and D.~{Trombetti},
  ``Distribution loss minimization by token ring control of power electronic
  interfaces in residential microgrids,'' \emph{IEEE Trans. Industrial
  Electronics}, vol.~59, no.~10, pp. 3817--3826, Oct. 2012.

\bibitem{tii}
R.~Bonetto, M.~Rossi, S.~Tomasin, and M.~Zorzi, ``On the interplay of
  distributed power loss reduction and communication in low voltage
  microgrids,'' \emph{IEEE Transactions on Industrial Informatics}, vol.~12,
  no.~1, pp. 322--337, Feb 2016.

\bibitem{seville}
R.~Bonetto, T.~Caldognetto, S.~Buso, M.~Rossi, S.~Tomasin, and P.~Tenti,
  ``Lightweight energy management of islanded operated microgrids for prosumer
  communities,'' in \emph{IEEE International Conference on Industrial
  Technology (ICIT)}, Seville, Spain, March 2015, pp. 1323--1328.

\bibitem{dr_1}
P.~B. Luh, L.~D. Michel, P.~Friedland, C.~Guan, and Y.~Wang, ``Load forecasting
  and demand response,'' in \emph{IEEE PES General Meeting}, Minneapolis, MN,
  U.S., July 2010.

\bibitem{add_1}
B.~Narayanaswamy, T.~S. Jayram, and V.~N. Yoong, ``Hedging strategies for
  renewable resource integration and uncertainty management in the smart
  grid,'' in \emph{IEEE PES Innovative Smart Grid Technologies Europe (ISGT
  Europe)}, Berlin,DE, Oct 2012.

\bibitem{add_2}
R.~Haque, T.~Jamal, M.~N.~I. Maruf, S.~Ferdous, and S.~F.~H. Priya, ``Smart
  management of phev and renewable energy sources for grid peak demand energy
  supply,'' in \emph{Electrical Engineering and Information Communication
  Technology (ICEEICT), 2015 International Conference on}, Dhaka, BD, May.

\bibitem{add_5}
G.~K. Venayagamoorthy, ``Potentials and promises of computational intelligence
  for smart grids,'' in \emph{2009 IEEE Power Energy Society General Meeting},
  Calgary, CA, July 2009.

\bibitem{svm}
V.~N. Vapnik, \emph{The Nature of Statistical Learning Theory}.\hskip 1em plus
  0.5em minus 0.4em\relax New York, NY, USA: Springer-Verlag New York, Inc.,
  1995.

\bibitem{svm_complexity}
L.~Bottou and C.-J. Lin, ``{Support Vector Machine Solvers},'' in \emph{Large
  Scale Kernel Machines}, L.~Bottou, O.~Chapelle, D.~{decoste}, and J.~Weston,
  Eds.\hskip 1em plus 0.5em minus 0.4em\relax Cambridge, MA, USA: MIT Press,
  2007, pp. 301--320.

\bibitem{ann_complexity}
P.~Orponen, ``Computational complexity of neural networks: A survey,''
  \emph{Nordic Journal of Computing}, vol.~1, no.~1, pp. 94--110, Mar. 1994.

\bibitem{for_1}
P.~Qingle and Z.~Min, ``Very short-term load forecasting based on neural
  network and rough set,'' in \emph{IEEE International Conference on
  Intelligent Computation Technology and Automation (ICICTA)}, Changsha, CN,
  May 2010.

\bibitem{svm_1}
K.~Gajowniczek and T.~Ząbkowski, ``Short term electricity forecasting using
  individual smart meter data,'' \emph{Procedia Computer Science}, vol.~35, pp.
  589 -- 597, 2014.

\bibitem{NNs}
S.~Haykin, \emph{Neural Networks: A Comprehensive Foundation}, 2nd~ed.\hskip
  1em plus 0.5em minus 0.4em\relax Upper Saddle River, NJ, USA: Prentice Hall
  PTR, 1998.

\bibitem{lstm}
S.~Hochreiter and J.~Schmidhuber, ``Long short-term memory,'' \emph{Neural
  Computation}, vol.~9, no.~8, Nov. 1997.

\bibitem{UCIDataSet}
\BIBentryALTinterwordspacing
K.~Bache and M.~Lichman, ``{UCI Machine Learning Repository},'' 2013. [Online].
  Available: \url{http://archive.ics.uci.edu/ml}
\BIBentrySTDinterwordspacing

\bibitem{bishop}
C.~M. Bishop, \emph{Neural Networks for Pattern Recognition}.\hskip 1em plus
  0.5em minus 0.4em\relax New York, NY, USA: Oxford University Press, Inc.,
  1995.

\bibitem{softsign}
X.~Glorot and Y.~Bengio, ``Understanding the difficulty of training deep
  feedforward neural networks,'' in \emph{Society for Artificial Intelligence
  and Statistics International Conference on Artificial Intelligence and
  Statistics (AISTATS’10)}, 2010.

\bibitem{adagrad}
J.~Duchi, E.~Hazan, and Y.~Singer, ``Adaptive subgradient methods for online
  learning and stochastic optimization,'' \emph{The Journal of Machine Learning
  Research}, vol.~12, Jul. 2011.

\end{thebibliography}
